\newif\ifarxiv
\newif\ifralfinal
\newif\ifconffinal
\ralfinalfalse \arxivfalse \conffinaltrue

\ifarxiv
\documentclass[letterpaper, 10 pt, conference]{ieeeconf}  %
\fi 
\ifralfinal
\documentclass[letterpaper, 10 pt, journal, twoside]{IEEEtran}
\fi
\ifconffinal
\documentclass[letterpaper, 10 pt, conference]{ieeeconf} %
\fi

\IEEEoverridecommandlockouts

\usepackage{fancyhdr}
\fancypagestyle{withfooter}{
  
  \fancyfoot[C]{\footnotesize Accepted to the IEEE ICRA Workshop On Neural Fields In Robotics (RoboNerf) 2024}
}

\usepackage{graphics} %
\usepackage{epsfig} %
\usepackage{times} %
\usepackage{amsmath} %
\usepackage{amssymb}  %
\usepackage{bbm}
\usepackage{booktabs}
\usepackage{mathtools}
\usepackage{multirow}
\usepackage{units}
\usepackage{bm}
\usepackage{xcolor}
\let\labelindent\relax
\usepackage{enumitem}
\usepackage{balance}
\usepackage[ruled,vlined]{algorithm2e}
\usepackage{algorithmicx}
\usepackage{cite} 
\usepackage[caption=false, font=footnotesize]{subfig}
\usepackage{graphicx}
\usepackage{adjustbox}
\usepackage{letltxmacro}
\LetLtxMacro{\originaleqref}{\eqref}
\renewcommand{\eqref}{Eq.~\originaleqref}
\usepackage{algpseudocode}
\algnewcommand\Algand{\textbf{and} }

\SetCommentSty{mycommfont} %
\usepackage{microtype}

\definecolor{LightCyan}{rgb}{0.88,1,0.88}
\definecolor{emb_color}{RGB}{252,224,225}
\definecolor{multi_head_attention_color}{RGB}{252,226,187}
\definecolor{add_norm_color}{RGB}{242,243,193}
\definecolor{ff_color}{RGB}{194,232,247}
\definecolor{softmax_color}{RGB}{203,231,207}
\definecolor{linear_color}{RGB}{220,223,240}
\definecolor{gray_bbox_color}{RGB}{243,243,244}

\ifarxiv
\usepackage{fancyhdr}
\fi
\makeatletter
\let\NAT@parse\undefined
\makeatother
\usepackage{hyperref}  %

\newcommand{\algoname}{Reg-NF}
\newcommand{\dataname}{ONR }
\begin{document}

\title{
\LARGE \bf

Object Registration in Neural Fields
}
\author{David Hall$^{1}$, Stephen Hausler$^{1}$, Sutharsan Mahendren$^{1,2}$, Peyman Moghadam$^{1,2}$   
\thanks{Reg-NF Website: \href{https://csiro-robotics.github.io/Reg-NF/}{https://csiro-robotics.github.io/Reg-NF}}
\thanks{$^1$ Authors are with the CSIRO Robotics, DATA61, CSIRO, Brisbane, QLD 4069, Australia. 
E-mails: {\tt\footnotesize \emph{firstname.lastname}@csiro.au}}
\thanks{
$^{2}$  Sutharsan Mahendren, and Peyman Moghadam are with the SAIVT research programme in the School of Electrical Engineering and Robotics, Queensland University of Technology (QUT), Brisbane, Australia.
E-mails: {\tt\footnotesize \emph\{sutharsan.mahendren, peyman.moghadam\}@qut.edu.au}
}
}

\bstctlcite{IEEEexample:BSTcontrol}

\maketitle
\ifarxiv
\thispagestyle{fancy}
\pagestyle{plain}
\fi

\thispagestyle{withfooter}
\pagestyle{withfooter}

\begin{abstract}
Neural fields provide a continuous scene representation of 3D geometry and appearance in a way which has great promise for robotics applications. 
One functionality that unlocks unique use-cases for neural fields in robotics is object 6-DoF registration.
In this paper, we provide an expanded analysis of the recent Reg-NF neural field registration method and its use-cases within a robotics context.
We showcase the scenario of determining the 6-DoF pose of known objects within a scene using scene and object neural field models.
We show how this may be used to better represent objects within imperfectly modelled scenes and generate new scenes by substituting object neural field models into the scene.

\end{abstract}

\section{Introduction}

For robotics applications, the six degree of freedom (6-DoF) registration between two scenes of interest is a crucial step, for tasks such as localisation, object pose estimation and 3D reconstruction. 
While many methods exist for representing 3D scenes, including point clouds, voxels and meshes, recently \emph{implicit representations} have emerged, which can compactly represent 3D scenes with unprecedented fidelity~\cite{barron2021mip, mildenhall2021nerf, wang2021neus, martin2021nerf, yariv2021volume, Fu2022GeoNeus, kobayashi2022distilledfeaturefields, muller2022instant}

Neural field (NF) registration is important for their use in robotic applications, as it enables uses such as the fusion of multiple implicit maps, and the ability to dynamically update an existing implicit field. 
Multiple works have looked to utilising NFs for registration~\cite{goli2023nerf2nerf, chen2023dreg, peat2022zero} but the first to look at direct registration of neural fields was nerf2nerf~\cite{goli2023nerf2nerf}. 
However, nerf2nerf does not suit the robotics domain as it relies on human-annotated keypoints for initialisation and assumes the scale of two neural fields are the same. 
Our recent work \algoname{}~\cite{hausler2024reg}, overcomes these limitations to estimate the relative 6-DoF pose transformation between two objects of interest which are located in two different neural fields. 
\algoname{} does not rely on human-annotated keypoints, operates directly on the continuous neural fields, and is capable of estimating transformation between two models with arbitrary scales. 
It builds upon the work on nerf2nerf, proposing a bidirectional registration loss, the use of multi-view sampling of the NF surface, and the use of SDFs~\cite{park2019deepsdf,chibane2020ndf, wang2021neus, Fu2022GeoNeus, yariv2021volume} as the implicit model of choice.
These increase registration accuracy, take advantage of NFs ability to render data from any view, and ensure a consistent and clear geometric representation of implicit models respectively.

\begin{figure}[t]
    \centering
    \includegraphics[width=\linewidth]{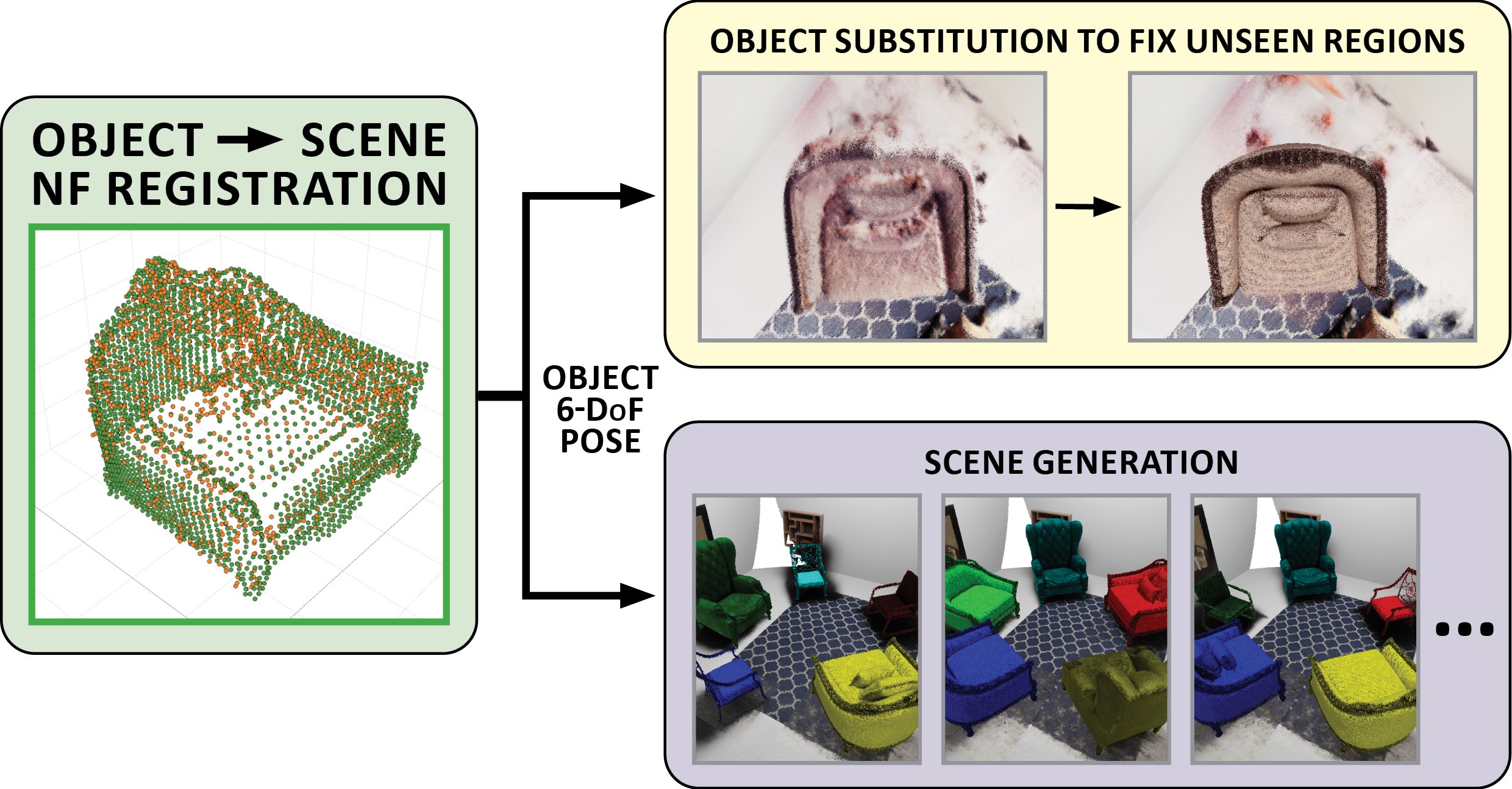}
    \vspace{-5mm}
    \caption{The use-case for object registration in neural fields examined in this paper. After an object neural field (NF) is registered with its counterpart in a scene and 6-DoF pose has been attained, the library NF can be substituted into the scene to fix errors in the scene NF or new scenes can be generated using other object NFs and the known object pose in the scene.}
    \label{fig:hero}
\vspace{-5mm}
\end{figure}

This paper is a companion to the original where we provide a condensed overview of our Reg-NF algorithm~\cite{hausler2024reg} and then extend the analysis of the original paper's results.
These consider the use-case scenario where \algoname{} is used to register a large scene NF with high-fidelity object-centric NFs stored in an object library, enabling both substitution of library objects into the scene, and replacement of object instances within the scene.
This provides two particular benefits for NF object registration in robotics highlighted in Fig.~\ref{fig:hero}. 
The first is object completion, using library substitution to improve the representation of scenes that have only partially observed objects.
The second is using instance replacement as a way to enable data-driven simulation where any scene NF can be edited (by replacing objects instances), generating new scene data for training in simulated NF environments. 

\section{Reg-NF}
\label{sec:method}

Reg-NF~\cite{hausler2024reg} provides a technique for aligning the surfaces of two different SDF NFs, by minimising the difference between their surfaces values.
This can be used to provide the 6-DoF pose of an object within a scene given both are represented as neural fields.
Here, we provide an overview of Reg-NF but invite readers to check the original paper~\cite{hausler2024reg} for full details.

\begin{figure}[t]
    \centering
    \includegraphics[width=0.9\linewidth]{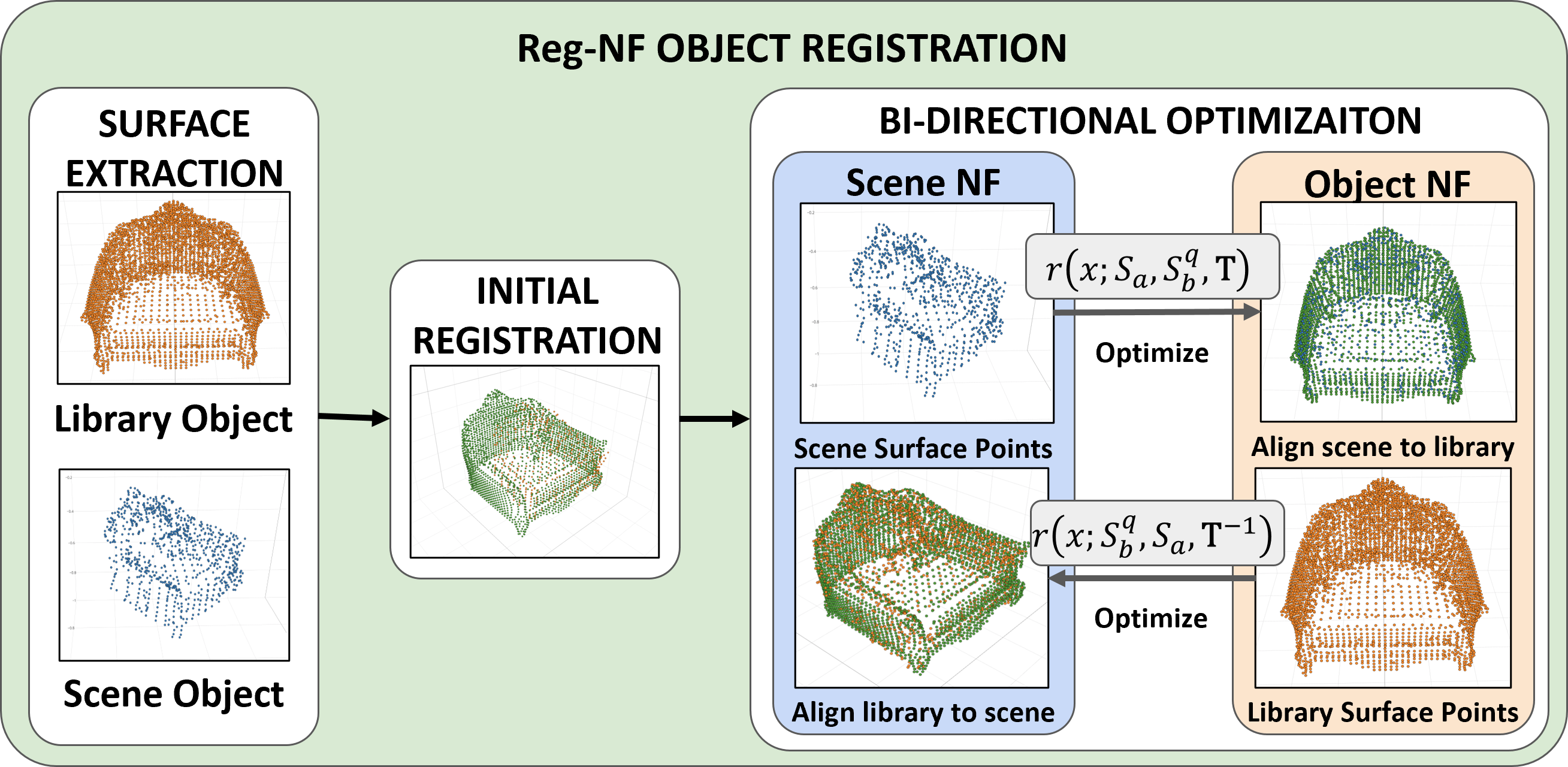}
    \vspace{-3mm}
    \caption{Overview of the \algoname{} registration process~\cite{hausler2024reg}. Blue and orange denotes surface sample points from the scene and library NFs for a matched object respectively. Green points represent the target alignment during optimisation. After surface extraction and an initial registration estimate, bi-directional optimisation iterates till convergence. Final output is a 6-DoF transformation matrix between models.}
    \label{fig:method}
\vspace{-2mm}
\end{figure}

For object registration, Reg-NF calculates the 6-DoF pose transformation $\textbf{T}$ of a detected object from a pre-existing NF object library within a larger scene NF.
This uses a differentiable optimisation function, initialised with an automated procedure. 
We denote $a$ as the notation of an implicit representation of a scene, and $b^q$ as the $q$th object-specific implicit model from a library of object neural fields where $q \in \{1, ... , Q\}$. 
For our work, we utilise volumetric implicit surface fields, specifically NeuS~\cite{wang2021neus} for all NFs in our work. 
These provide colour $c(\textbf{x},\textbf{v})$ and signed surface distance $S(\textbf{x})$ mapping functions derived from a shared backbone network with separate SDF and colour output heads when given a 3D point $\textbf{x}$ and viewing direction $\textbf{v}$. 
For full details on NeuS, please refer to~\cite{wang2021neus}.
An overview of Reg-NF is provided in Fig.~\ref{fig:method}.

\subsection{Initial Registration}

\begin{figure}
    \centering
    \includegraphics[width=0.45\linewidth]{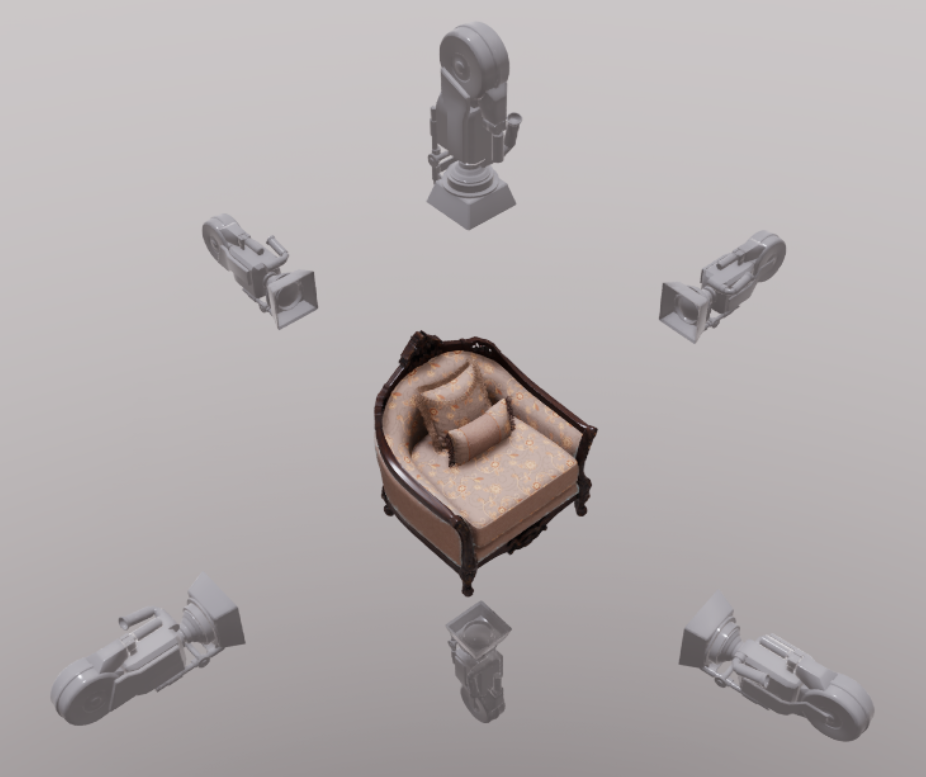}
    \caption{Example of multiple views used for surface point extraction.}
    \label{fig:multi-cam}
    \vspace{-7mm}
\end{figure}

\algoname{} begins by establishing approximate correspondences between objects of interest within $a$ and $b^q$. 
Assuming an object detection provides approximate location and classification of an object in $a$ that matches to $b^q$, we calculate a set of $N$ camera views pointed at the centroid of the detected object of interest as shown in Fig~\ref{fig:multi-cam}. 
We generate a grid pattern of rays travelling from each camera pose and return surface sample points $P_a$ and $P_b^q$ for our two respective neural fields.
This multi-view sampling approach provides a clear geometric representation of the object for initialisation and leaves it less susceptible to poor surface sample initialisation from a single poor camera view.

Once surface samples are found for both NFs, we employ RANSAC~\cite{fischler1981random} with Fast Point Feature Histogram (FPFH)~\cite{rusu2009fast} descriptors to estimate the correspondence between source and target, further refined via the point-to-point Iterative Closest Point (ICP)~\cite{121791} method. 
Through this, we attain the initial six-degree-of-freedom (6-DoF) pose transformation $\hat{\textbf{T}}$. Our initial registration takes $\approx 14$ seconds on average.

\subsection{Bi-directional Optimization}

Reg-NF treats registration as an optimisation function that uses gradient decent to find the optimal pose transformation $\textbf{T}$ between two surface fields initialised using $\hat{\textbf{T}}$, and assuming an initial scale factor of $s = 1$ that minimizes a loss function $L_s$.  
Building on nerf2nerf~\cite{goli2023nerf2nerf}, optimisation is performed over a discrete set of samples, with a robust kernel $\kappa$ with learnt parameters $p$ and $\alpha$ used to improve the robustness against outlier samples. 
However, Reg-NF improves optimisation by collecting surface samples points from both $a$ and $b^q$ ($A$ and $B^q$), and performing a bidirectional optimisation over the model SDF outputs ($S_a$ and $S_b^q$), as well as by including a regulariser designed to penalise when $A$ and $B^q$ deviate after transformation.
Given this, the loss function is expressed as :
\begin{equation}
\begin{split}
    L_s(S_a, S_b^q; \textbf{T}) = E_{x \in A} \kappa (r(x; S_a, S_b^q, \textbf{T}); p, \alpha) \\
    + E_{x \in B^q} \kappa (r(x; S_b^q, S_a, \textbf{T}^{-1}); p, \alpha) 
    + w L_r ,
\end{split}
\end{equation}
where registration residuals $r$ are expressed as:
\begin{equation}
    r(x; S_a, S_b^q, \textbf{T}) = \| S_a - S_b^q\textbf{T} \| \quad x \in A ,
\end{equation}
and 
\begin{equation}
    r(x; S_b^q, S_a, \textbf{T}^{-1}) = \| S_b^q - S_a\textbf{T}^{-1} \| \quad x \in B^q,
\end{equation}
and $L_r$ is our regulariser weighted with weight factor $w$ expressed as:
\begin{equation}
    L_r = \sum \frac{\min_{B^q} D_{A,B^q}^2}{\vert A \vert}, D_{A,B^q}^2 = \| A - B^q\textbf{T} \|^2,
\end{equation}
where $\vert A \vert$ denotes the number of samples in $A$.
Our loss function $L_s$ is calculated using the current estimated 6-DoF pose transform $\textbf{T}$ composed from translation components along all axes $[t_x, t_y, t_z]$, roll, pitch and yaw Euler angles $[r_r,r_p,r_y]$, and a scaling factor $\sigma$  applied to all axes. 
Our loss function is then optimised over the learnt parameters: $(t_x,t_y,t_z,r_r,r_p,r_y,s,p,\alpha)$. Our optimization procedure takes $\approx 16$ seconds on average.

\begin{figure}[t]
    \captionsetup[subfigure]{labelformat=empty}
    \centering
    \subfloat[end table (\textit{et})]{\includegraphics[width=0.2\linewidth]{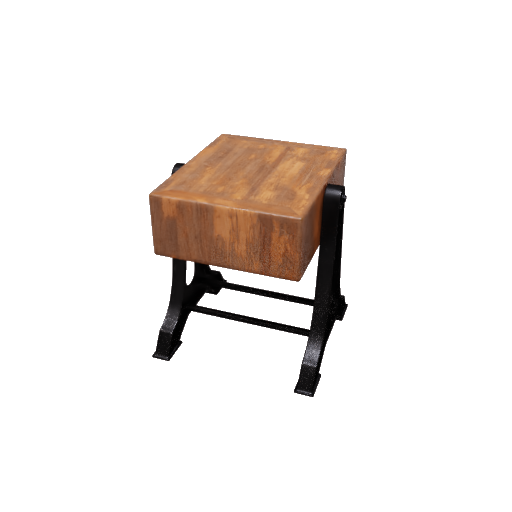}}
    \quad
    \subfloat[table (\textit{t})]{\includegraphics[width=0.2\linewidth]{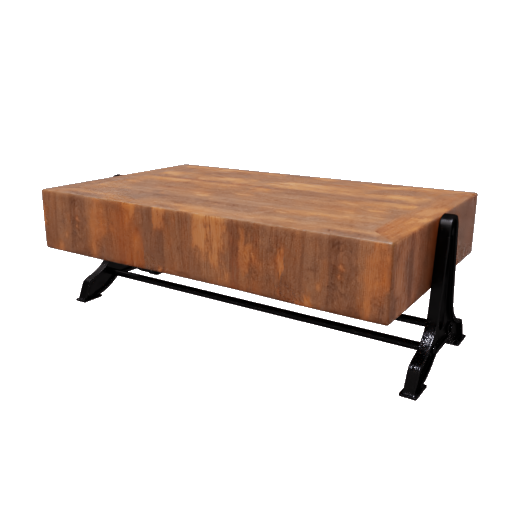}}
    \quad
    \subfloat[willow table (\textit{wt})]{\includegraphics[width=0.22\linewidth]{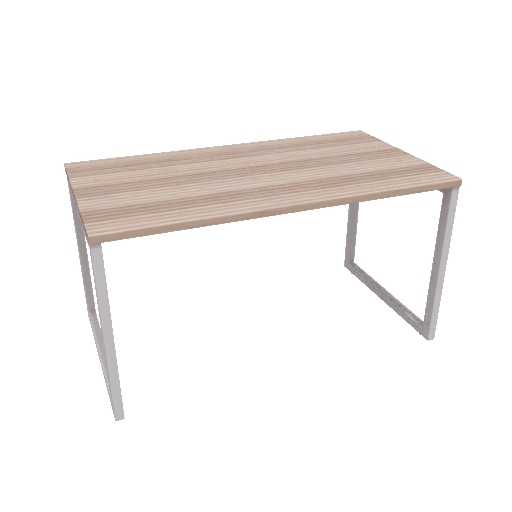}}
    \quad
    \subfloat[chair (\textit{c})]{\includegraphics[width=0.2\linewidth]{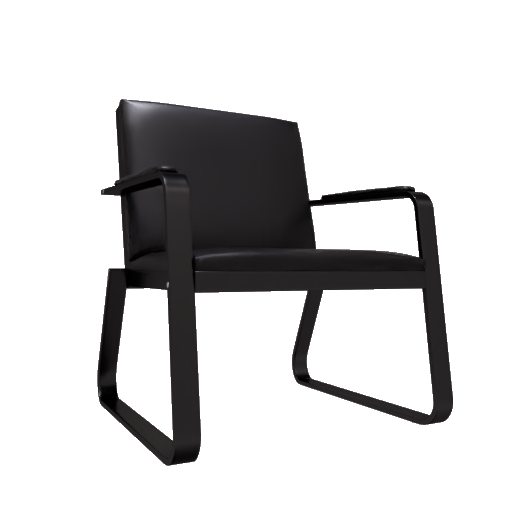}}
    \quad
    \subfloat[fancy chair (\textit{fc})]{\includegraphics[width=0.2\linewidth]{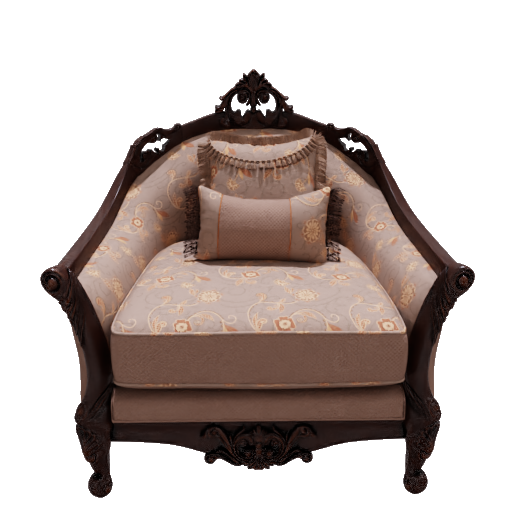}}
    \quad
    \subfloat[dining chair (\textit{dc})]{\includegraphics[width=0.22\linewidth]{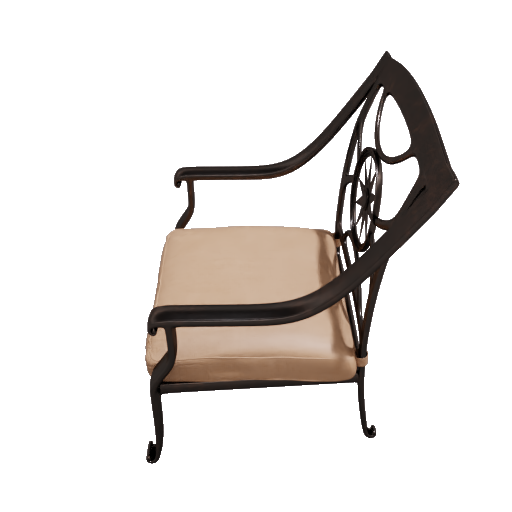}}
    \quad
    \subfloat[fancy chair w/o pillow (\textit{fc-nop})]{\includegraphics[width=0.2\linewidth]{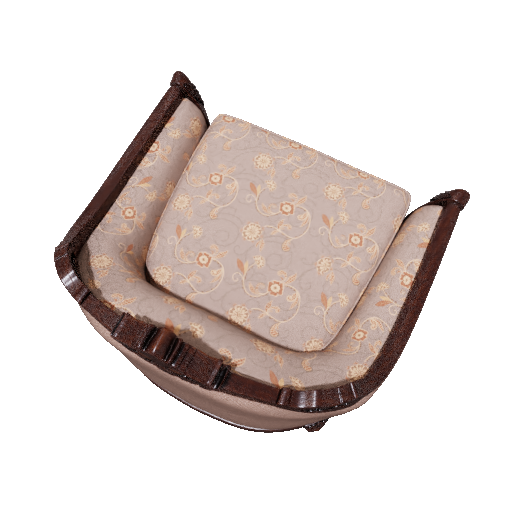}}
    \quad
    \subfloat[matrix chair (\textit{mc})]{\includegraphics[width=0.24\linewidth]{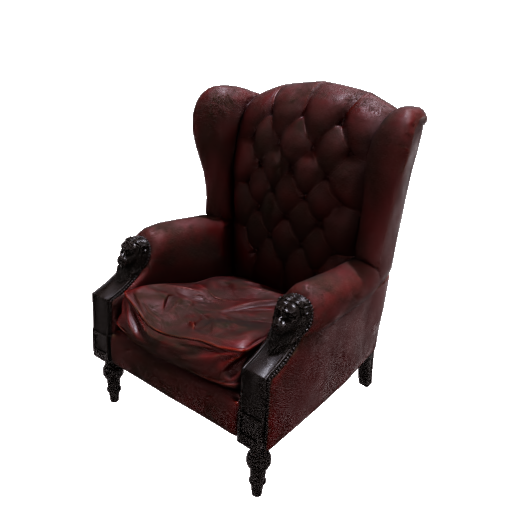}}
    \caption{Example images of object model images in ONR dataset.}
    \label{fig:onr_objects}
    \vspace{-6mm}
\end{figure}

\begin{table*}[t!]
\centering
\caption{Comparison between \algoname{} other registration methods.}
\label{tbl:n2n_compare}
\begin{tabular}{cc|ccccc|ccc|ccc}
 &               & \multicolumn{5}{c|}{\textbf{All Chairs Room}}                                      & \multicolumn{3}{c|}{\textbf{All Tables Room}} & \multicolumn{3}{c}{\textbf{Mix Room}}             \\ \cline{3-13} 
  &               & c              & dc             & fc             & fc-nop         & mc             & et            & t            & wt             & dc             & fc             & t              \\ \hline
\multicolumn{1}{c|}{\multirow{4}{*}{\textbf{$ \Delta \mathbf{t}\downarrow$}}}                   & FGR~\cite{zhou2016fast} & 0.868 & 1.084 & 0.3715 & 0.485 & 0.152 & 0.368 & 0.819 & 0.219 & 1.351 & 0.226 & 0.636\\
\multicolumn{1}{c|}{}  & nerf2nerf~\cite{goli2023nerf2nerf}           & 0.278          & 0.291          & 0.127          & 0.117          & \textbf{0.0007}          & 0.525              & 0.084            & 0.106               & 0.343          & 0.154          & 0.086          \\
\multicolumn{1}{c|}{} & \algoname-init (ours) & 0.300 & 0.322 & 0.093 & 0.110 & 0.125 & 0.383           & 0.210          & 0.144          & 0.447 & 0.138 & 0.100 \\
\multicolumn{1}{c|}{}                                                         & \textbf{\algoname} (ours) & \textbf{0.04} & \textbf{0.035} & \textbf{0.018} & \textbf{0.014} & 0.007 & \textbf{0.398}           & \textbf{0.044}          & \textbf{0.022}          & \textbf{0.292} & \textbf{0.029} & \textbf{0.009} \\ \hline
\multicolumn{1}{c|}{\multirow{4}{*}{\textbf{$ \Delta \mathbf{R}\downarrow$}}} & FGR~\cite{zhou2016fast} & 1.130 & 1.669 & 1.020 & 1.449 & 0.1389 & 2.513 & 1.656 & 1.282 & 1.669 & 0.395 & 2.297\\
\multicolumn{1}{c|}{}  & nerf2nerf~\cite{goli2023nerf2nerf}           & \textbf{0.041}          & 0.088          & 0.050          & 0.034 & \textbf{0.002} & \textbf{2.396}              & \textbf{0.002}             & 0.221               & \textbf{0.050} & 0.034          & 0.026          \\
\multicolumn{1}{c|}{} & \algoname-init (ours) & 0.213 & 0.196 & 0.044 & 0.074 & 0.032 & 2.566           & 0.235          & 0.140          & 0.195 & 0.099 & 0.016 \\
\multicolumn{1}{c|}{}                                                         & \textbf{\algoname} (ours) & 0.048 & \textbf{0.039} & \textbf{0.031} & \textbf{0.020}          & 0.009          & 2.6           &    0.053         & \textbf{0.025}          & 0.641          & \textbf{0.030} & \textbf{0.012} \\ \hline
\multicolumn{1}{c|}{\multirow{4}{*}{\textbf{$ \Delta s \downarrow$}}} &    FGR~\cite{zhou2016fast} & NA & NA & NA & NA & NA & NA & NA & NA & NA & NA & NA\\
\multicolumn{1}{c|}{}  & nerf2nerf~\cite{goli2023nerf2nerf}       & NA       & NA       & NA       & NA        & NA    & NA              & NA              & NA           & NA          & NA          & NA          \\
\multicolumn{1}{c|}{} & \algoname-init (ours) & NA & NA & NA & NA & NA & NA           & NA          & NA          & NA & NA & NA \\
\multicolumn{1}{c|}{} &
\textbf{\algoname} (ours)     &    \textbf{0.019}     &  \textbf{0.007}      &  \textbf{0.007}      &    \textbf{0.060}      &     \textbf{0.003}  &      \textbf{0.019}           &       \textbf{0.004}          &       \textbf{0.006}      &      \textbf{0.021}       &        \textbf{0.009}     &      \textbf{0.005}       \\
 \hline
\end{tabular}
\vspace{-6mm}
\end{table*}

\section{Experimental Design}

\begin{figure}[t]
    \centering
    \subfloat[ac-room]{\includegraphics[width=0.25\linewidth]{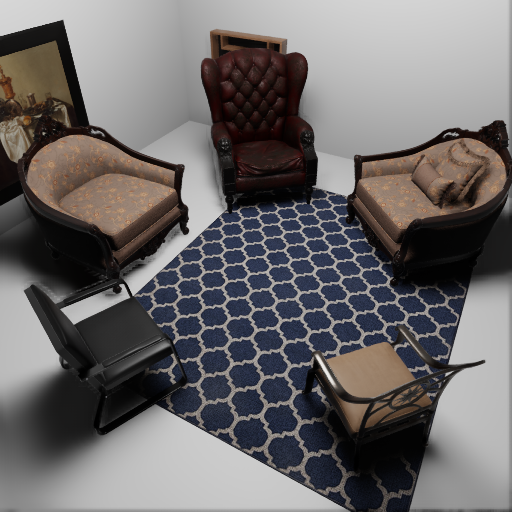}}
    \quad
    \subfloat[atbl-room]{\includegraphics[width=0.25\linewidth]{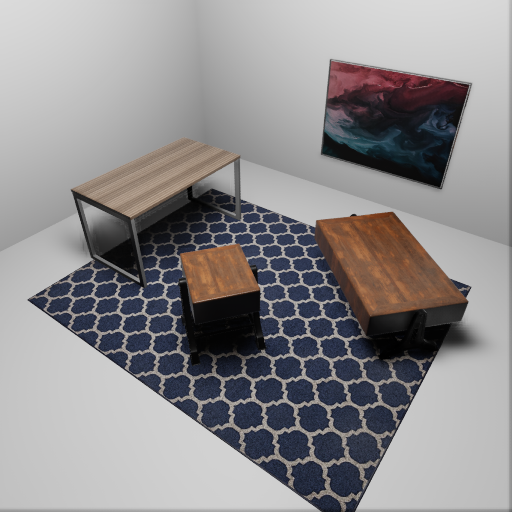}}
    \quad
    \subfloat[mix-room]{\includegraphics[width=0.25\linewidth]{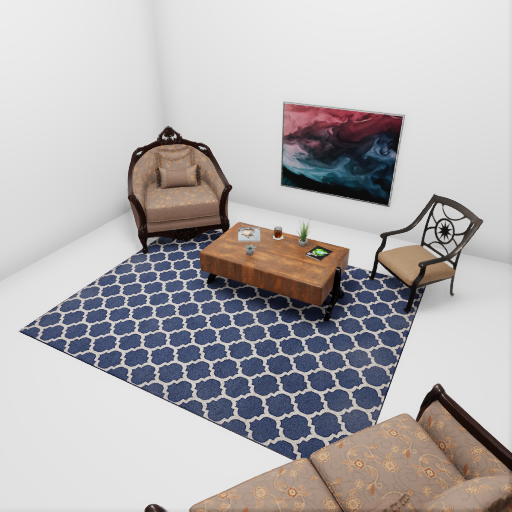}}
    \caption{Scene models in \dataname dataset.}
    \label{fig:datasets}
    \vspace{-8mm}
\end{figure}

\noindent\textbf{Dataset}: The dataset we use for our experiment comprises high-fidelity simulated images and corresponding camera poses of objects and scenes, collected using NVIDIA's Omniverse Isaac Sim platform.
We will refer to this dataset as our Object NF Registration (ONR) dataset.
Object data is of single objects in a ``void'' and scene data is a standardised room with different object models placed within.
Object data becomes the basis for our object NF library and contains data for 5 chair and 3 table models learnt at varying scales to maximize fidelity.
ONR objects are outlined in Fig~\ref{fig:onr_objects}
There are three scenes collected: 1) containing all chair models stored as object data (ac-room); 2) containing all table models stored as object data (atbl-room); 3) a scene with a mixture of two chairs and a cluttered table (mix-room). 
Example images from each scene are shown in Fig.~\ref{fig:datasets}.

\begin{table*}[]
\caption{Single view vs Multi-view Tests.}
\label{tbl:single-vs-multi}
\centering
\begin{tabular}{cc|cccccccc|}
\multicolumn{2}{c|}{}                                                                             & \textbf{c} & \textbf{dc} & \textbf{fc} & \textbf{fc-nop} & \textbf{mc} & \textbf{et} & \textbf{t} & \textbf{wt} \\ \hline
\multirow{2}{*}{\textbf{$\Delta \mathbf{t}\downarrow$}} & \textbf{Single}        & 0.307      & 0.674       & 0.378       & 0.681           & 0.147       & 0.449       & 0.632      & 0.733       \\
                                                                         & \textbf{Multi. (ours)} & \textbf{0.040}      & \textbf{0.035}       & \textbf{0.018}       & \textbf{0.014}           & \textbf{0.007}       & \textbf{0.398}       & \textbf{0.044}     & \textbf{0.022}       \\ \hline
\multirow{2}{*}{\textbf{$ \Delta \mathbf{R}\downarrow$}}                 & \textbf{Single}        & 0.931      & 2.201       & 2.182       & 0.842           & 0.144       & \textbf{1.562}       & 0.731      & 2.527       \\
                                                                         & \textbf{Multi. (ours)} & \textbf{0.048}      & \textbf{0.039}       & \textbf{0.031}       & \textbf{0.020}      & \textbf{0.009}       & 2.600       & \textbf{0.053}      & \textbf{0.025}       \\ \hline
\multirow{2}{*}{\textbf{$ \Delta s \downarrow$}} & \textbf{Single}        & 0.044      & 0.047       & 0.498       & 0.339           & 0.178       & 0.121       & 0.470      & 0.060       \\
                                                                         & \textbf{Multi. (ours)} & \textbf{0.019}      & \textbf{0.007}       & \textbf{0.007}       & \textbf{0.060}           & \textbf{0.003}       & \textbf{0.019}       & \textbf{0.004}      & \textbf{0.006}       \\ \hline
\end{tabular}
\vspace{-8mm}
\end{table*}

\noindent\textbf{Metrics}: We follow the metrics in nerf2nerf~\cite{goli2023nerf2nerf} for our work.
We report the root mean squared error (RMSE) between the ground-truth and predicted scene to object transformation matrices.
Rotation error $\Delta \mathbf{R}$ is calculated in radians and translation error $\Delta \mathbf{t}$ is calculated in normalised object frame units.
We also report the absolute difference between estimated and true scale between scene and object models $\Delta s$.
Note that this is given instead of RMSE as scale factor is assumed consistent across all axes.

\noindent\textbf{Nerf2nerf on ONR dataset}: To evaluate nerf2nerf on the ONR dataset, we enabled nerf2nerf to utilise the same surface fields from our SDF models as used by \algoname{}. 
We manually generated new human annotated keypoints for the initialisation procedure used in nerf2nerf. 

\noindent\textbf{Training models}: All NF models were trained using the sdfstudio~\cite{Yu2022SDFStudio} implementation of NeuS which includes the proposal network from MipNeRF-360~\cite{barron2022mip} for training speed-up (neus-facto). For more details please refer to sdfstudio~\cite{Yu2022SDFStudio}.

\noindent\textbf{Object Proposals}: \algoname{} assumes a detection has already been made within a scene's NF through some pre-existing method.
As object proposal generation is not within the scope of this work, we utilise ground-truth object 3D bounding boxes to calculate initial set of $N$ camera extrinsics for generating the initial surface samples. 

\noindent\textbf{\algoname{} hyperparameters}: We provide the following hyperparameters for Reg-NF. For our sampler, we use $\omega_1=0.01$, $\omega_2 = 0.02$ and $\xi = 0.02$. We set $\rho$ to $r/20$, where $r$ is the scene radii and generate new samples every $10$ iterations. We use a learning rate of $0.02$ for rotation, $0.01$ for translation, $0.01$ for scale, and $0.005$ for adaptive kernel parameters, for a maximum of $200$ iterations. We also have early stopping criteria, when $\sum (r(x; S_a, S_b^q, \textbf{T}), \, x \forall A)/\vert A \vert \leq 0.0005$.

\section{Results}
We first perform a quantitative analysis of \algoname{}, comparing it to nerf2nerf, FGR, and the output of our initialisation step to demonstrate that we are outperforming them while not requiring manually annotated keypoints or an assumption that all objects are of the same scale as the scene.
This is followed by experiments  showing the benefit of \algoname{} multi-view surface extraction. Finally, we demonstrate the benefits of \algoname{} for modelling imperfect scene NFs with known object NF replacement, and show how \algoname{} can enable object instance replacement for generating alternative NF scenes with the same underlying object arrangements but different object NF models.

\subsection{Comparison to nerf2nerf}

\begin{figure}[t]
    \centering
    \subfloat[nerf2nerf~\cite{goli2023nerf2nerf}]{\includegraphics[width=0.35\linewidth]{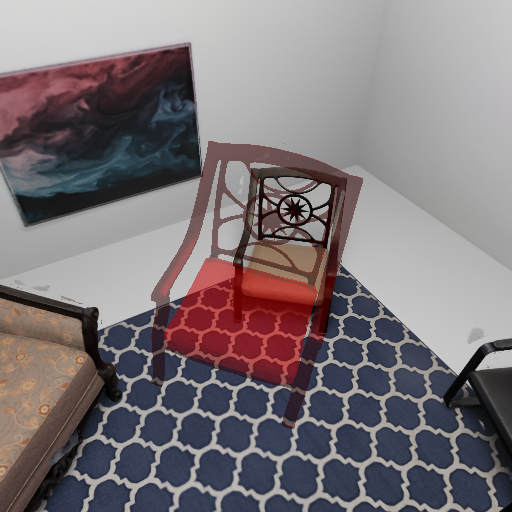}}
    \quad
    \subfloat[Reg-NF~\cite{hausler2024reg}]{\includegraphics[width=0.35\linewidth]{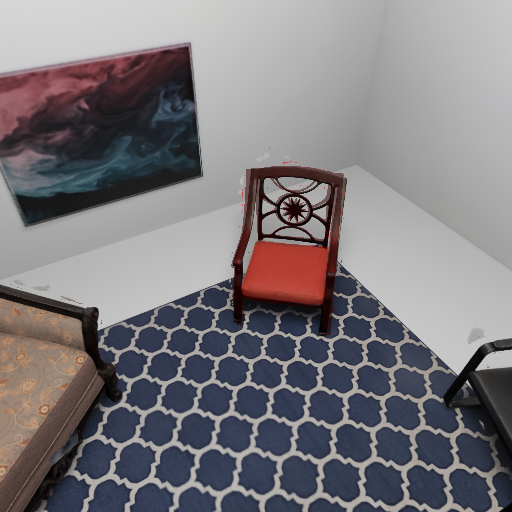}}
    \caption{Qualitative comparison of NF object registration methods for \textit{dc} library model (red) in \textit{ac-room} scene.
    Scale of \textit{dc} in library starts much larger than within the scene causing nerf2nerf~\cite{goli2023nerf2nerf} to fail.}
    \label{fig:n2n_comparison}
    \vspace{-8mm}
\end{figure}

We evaluate and compare the performance of \algoname{}, nerf2nerf~\cite{goli2023nerf2nerf}, FGR~\cite{zhou2016fast}, and our Reg-NF initialisation on our \dataname dataset. 
In Table~\ref{tbl:n2n_compare} we see that \algoname{} is typically at least an order of magnitude better than nerf2nerf in terms of $\Delta \textbf{t}$ and is still generally superior in $\Delta \textbf{R}$. We attribute the large increase of errors for nerf2nerf as being primarily due to the inherent scale differences between scene and database object models, for which nerf2nerf has no functionality to handle. 
An example of a nerf2nerf failure case is shown in Fig.~\ref{fig:n2n_comparison}

\begin{figure}
    \centering
    
    \subfloat{\includegraphics[width=0.25\linewidth]{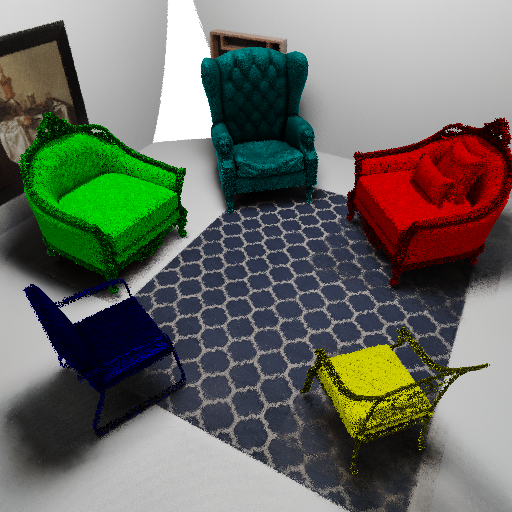}}
    \quad
    \subfloat{\includegraphics[width=0.25\linewidth]{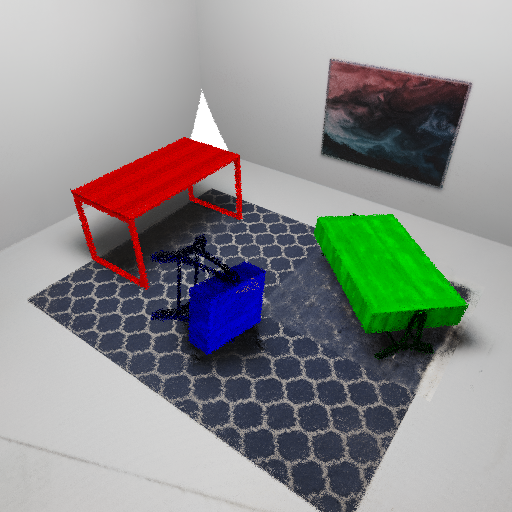}}
    \quad
    \subfloat{\includegraphics[width=0.25\linewidth]{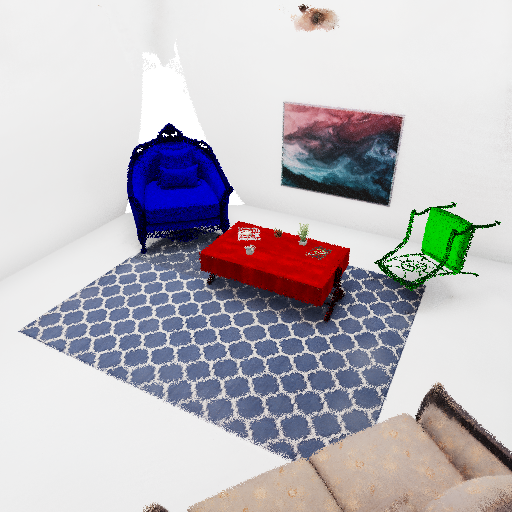}}
    \caption{Example of library substitution using \algoname{} for all objects shown in Fig~\ref{fig:datasets}. Substitutions are based on \algoname{} outputs. Colours are added to object NFs during render to provide visual distinction.}
    \label{fig:all-scene-replace}
    \vspace{-7mm}
\end{figure}

We also observe that despite the Reg-NF initialisation providing improved results over FGR, that the initialisation is rough and requires Reg-NF optimization to provide a close-fitting registration.
Focusing on \algoname{}, we note that failures can still occur, such as when we match object \emph{dc} to scene \emph{mix Room} or \emph{et} to scene \emph{at-room}.
In both these cases, we note the cause of failure being poor initialisation that proved inescapable for \algoname{}, even if the error metrics of initialisation are lower than for some that were able to improve (e.g. \textit{c} in \textit{ac-room}).
A qualitative analysis of the \algoname{} registration can be seen in Fig~\ref{fig:all-scene-replace}.

\subsection{Effect of multi-view sample initialisation}
\begin{figure}
    \centering
    \subfloat{\includegraphics[width=0.2\linewidth]{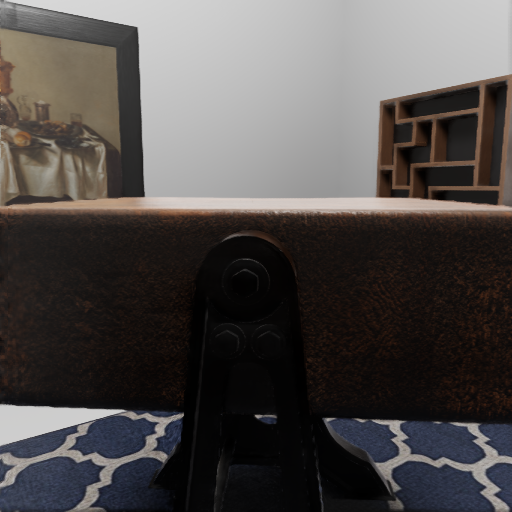}}
    \quad
    \subfloat{\includegraphics[width=0.2\linewidth]{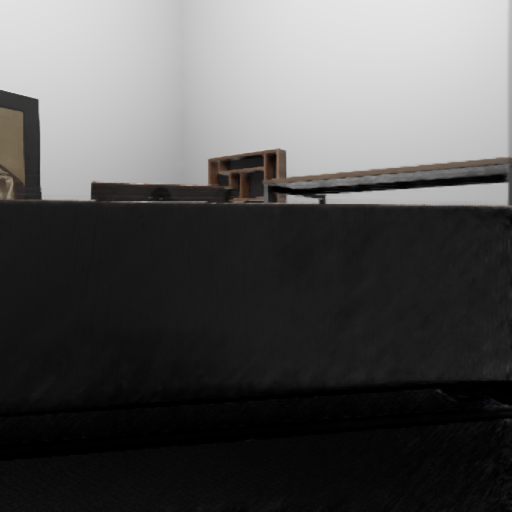}}
    \quad
    \subfloat{\includegraphics[width=0.2\linewidth]{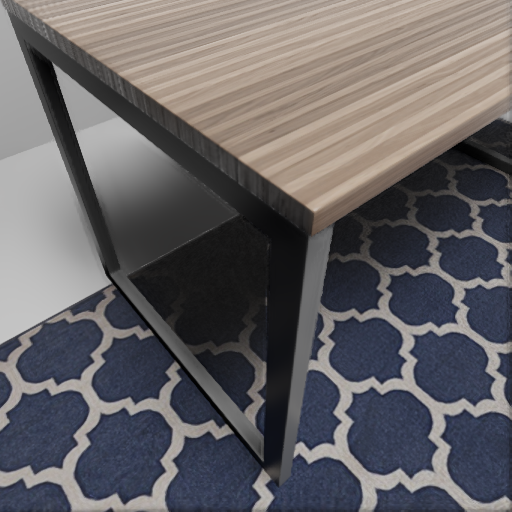}}
    \quad
    \subfloat{\includegraphics[width=0.2\linewidth]{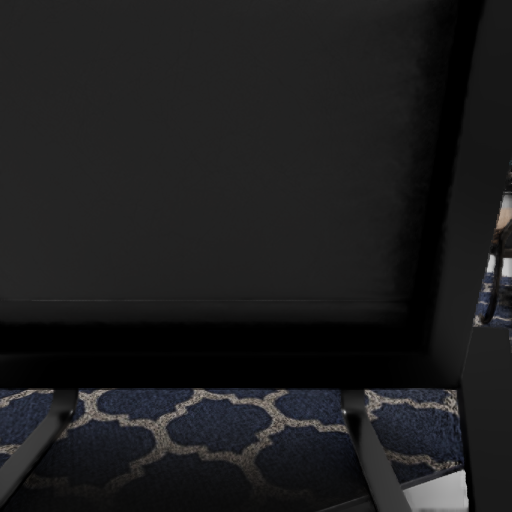}} 
    \quad
    \subfloat{\includegraphics[width=0.2\linewidth]{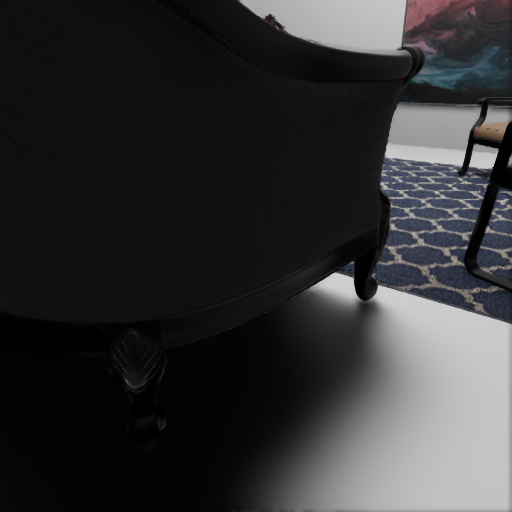}}
    \quad
    \subfloat{\includegraphics[width=0.2\linewidth]{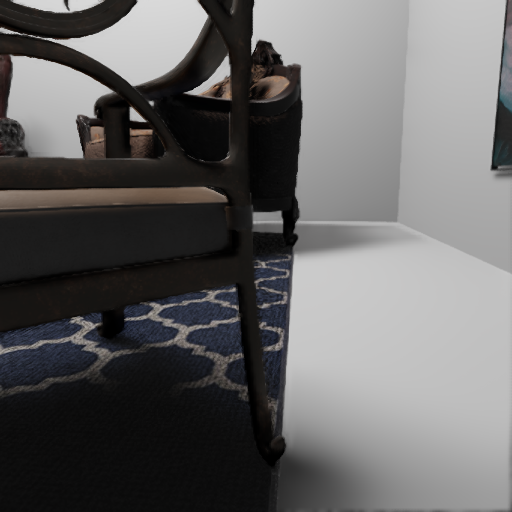}}
    \quad
    \subfloat{\includegraphics[width=0.2\linewidth]{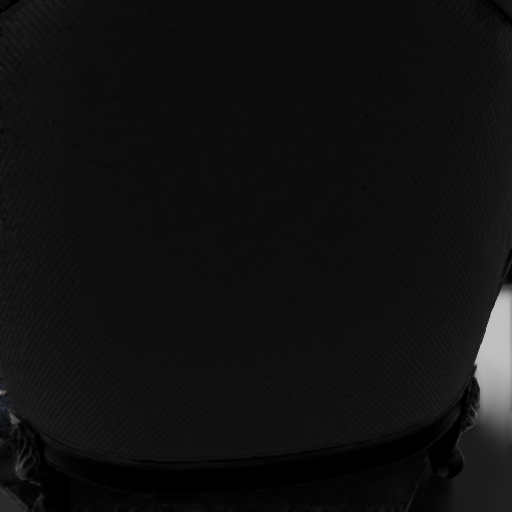}}    
    \quad
    \subfloat{\includegraphics[width=0.2\linewidth]{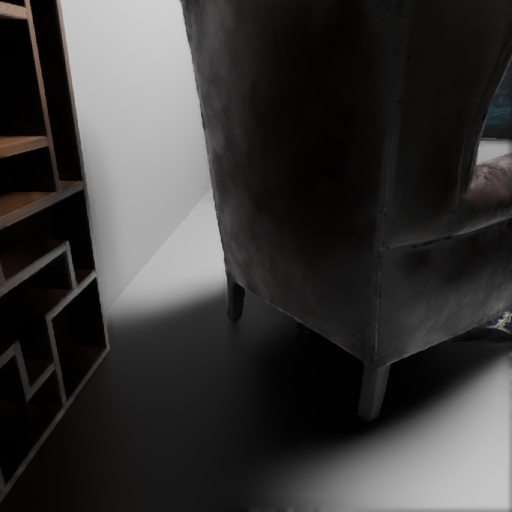}}
    
    \caption{Highly uninformative and/or partial views of models in ac-room and at-room from training data used for single-view experiments. 
    In order: \textit{et}, \textit{t}, \textit{wt}, \textit{c}, \textit{fc}, \textit{dc}, \textit{fc-nop} and \textit{mc} objects. 
    }
    \label{fig:hard-single-views}
    \vspace{-7mm}
\end{figure}

The benefit of multi-view sampling during initialisation is most felt when an object has no distinguishing characteristics or is only partially seen from a single viewing angle taken from the training data.
Using a single view introduces a high level of variability in object coverage.
Using views shown in Fig~\ref{fig:hard-single-views} from neural field training data that only view part of the desired object for sampling the surface field, we test the worst-case scenario of single-view experiments.
The rest of the Reg-NF pipeline is kept consistent and quantitative results are shown in Table~\ref{tbl:single-vs-multi}.
This shows using bad viewpoints drastically reduces performance as no meaningful features could be extracted to enable effective registration.

\subsection{Substitution within imperfect scene models}
\begin{figure}[t]
    \centering
    \subfloat[Original scene]{\includegraphics[width=0.3\linewidth]{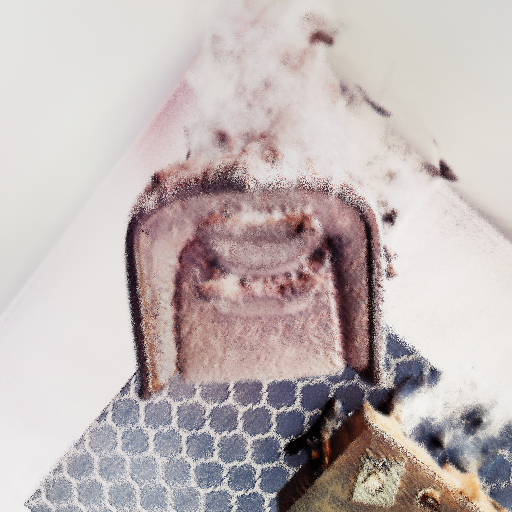}}
    \quad
    \subfloat[Library substitution]{\includegraphics[width=0.3\linewidth]{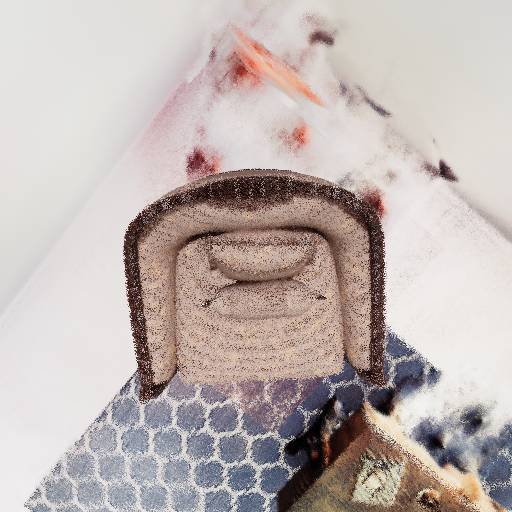}}
    \caption{Example of library substitution in a scene with low coverage. Original NF (a) cannot correctly render regions unseen during training. (b) library substitution after \algoname{} registration. Geometry of the object within the scene can be fully rendered from only partial initial view.}
    \label{fig:use-case-short}
    \vspace{-5mm}
\end{figure}

To demonstrate practical applications for substitution using \algoname{}, we consider when a robot may not be able to fully traverse a scene to get ``full coverage'' of an object for the scene's NF.
We can see in Fig.~\ref{fig:use-case-short} that a scene NF trained from a low-coverage trajectory cannot render the back of the chair clearly as it never saw that during training.
Using \algoname{}, we register the object NF for the imperfect chair within the scene and substitute it in the scene to render a clear view of the back of the chair.

\subsection{Instance replacement for scene generation}
\begin{figure}[t]
    \centering
    \subfloat{\includegraphics[width=0.2\linewidth]{Figures/ac_full_replace.png}}
    \quad
    \subfloat{\includegraphics[width=0.2\linewidth]{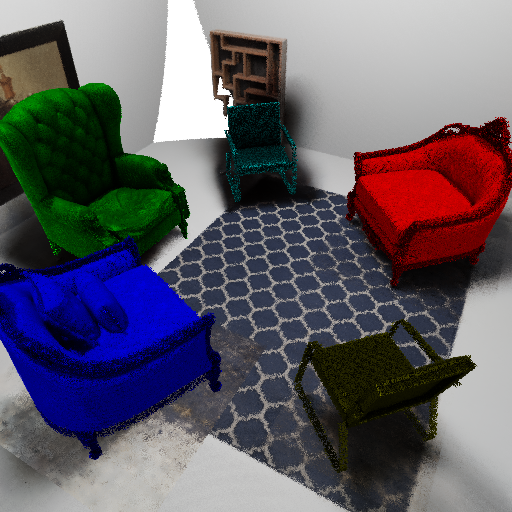}}
    \quad
    \subfloat{\includegraphics[width=0.2\linewidth]{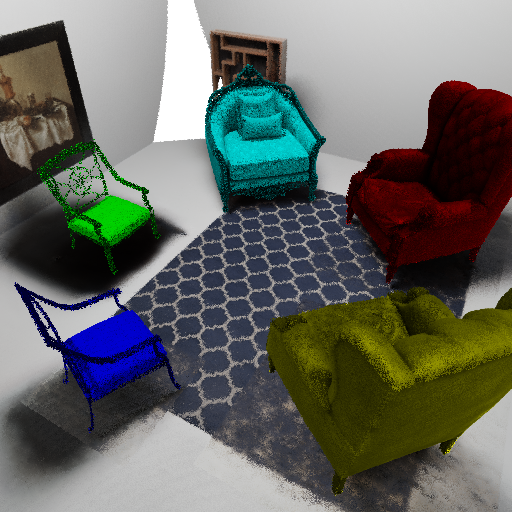}}
    \quad
    \subfloat{\includegraphics[width=0.2\linewidth]{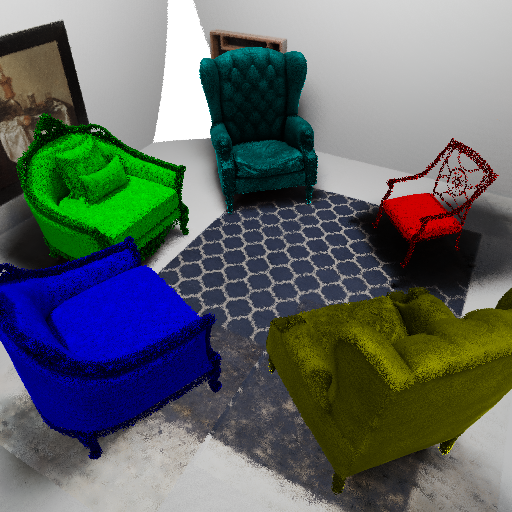}}
    \caption{Example of object instance replacement for generating new scenes. Original layout with coloured substitutions (left) followed by random replacement of chair models with others in the model library. }
    \label{fig:instance replacement}
    \vspace{-7mm}
\end{figure}

Finally, we demonstrate the benefits of using a library of pre-trained NF objects for creating new scenes.
Once \algoname{} derives the transform between a matched object NF and the scene NF, the known relative shapes/poses of objects within the NF library of the same class can be used to replace registered scene objects with any other object instance.
We demonstrate this in Fig.~\ref{fig:instance replacement} showing where, after a scene has been turned into a neural field, objects within that scene can be changed to provide new data based on the layout of the original scene.

\section{Conclusions}
This paper provides an extended analysis of \algoname{}, a novel method for registration between neural field (NF) representations.
Specifically, we examine the scenario where Reg-NF calculates the 6-DoF transform between objects found in a scene NF and object-centric NF counterparts stored in an NF object library.
We analyse the effectiveness of \algoname{}, showcasing how it's bi-directional optimization improves upon initial registration, and how it's multi-view surface sampling provides robustness against naive single-view sampling.
We demonstrate its advantages for modelling objects within imperfect scene NFs and for enabling data-driven robotics research by generating modified scene NFs for robots to train in.

\balance{}

\bibliographystyle{IEEEtran}
\bibliography{refs}

\end{document}